%% file: main.tex
\documentclass{l4dc2026}
\title[Warm-starting active-set solvers using graph neural networks]{Warm-starting active-set solvers using graph neural networks}
\usepackage{times}
\usepackage{xcolor}                                     % colors
\usepackage{enumitem}                                   % enumeration
\usepackage{mathtools}                                  % coloneqq
\usepackage{graphicx,tikz,caption,subcaption}           % figures 
\usepackage{hyperref}                                   % links
\usepackage{comment}                                    % comments
\usepackage{booktabs}                                   % tables
\usepackage{multirow}                                   % multicolumn/row
\usepackage{tikz}                                       % create figure
\usepackage{wrapfig}                                    % wrap figure
\usetikzlibrary{shapes.geometric, arrows, positioning,fit,calc}
\usepackage{algorithm}
\usepackage{algorithmic}                                % algorithmicx style
\newcommand{\algcomment}[1]{\hfill \texttt{// #1}}

\definecolor{lightblue}{HTML}{A6CEE3}
\definecolor{darkblue}{HTML}{1F78B4}
\definecolor{lightgreen}{HTML}{B2DF8A}
\definecolor{darkgreen}{HTML}{33A02C}
\definecolor{lightred}{HTML}{FB9A99}
\definecolor{darkred}{HTML}{E31A1C}
\definecolor{lightorange}{HTML}{FDBF6F}
\definecolor{darkorange}{HTML}{FF7F00}
\definecolor{lightpurple}{HTML}{CAB2D6}
\definecolor{darkpurple}{HTML}{6A3D9A}

\SetCommentSty{\normaltext}
\usepackage{natbib}
\setcitestyle{authoryear,open={(},close={)}}

\coltauthor{\Name{Ella J. Schmidtobreick} \Email{ella-johanna.schmidtobreick@it.uu.se}\\
 \Name{Daniel Arnström} \Email{daniel.arnstrom@it.uu.se}\\
 \Name{Paul Häusner} \Email{paul.hausner@it.uu.se}\\
 \Name{Jens Sjölund} \Email{jens.sjolund@it.uu.se}\\
 \addr Department of Information Technology, Uppsala University, Sweden}

\begin{document}

\maketitle

\begin{abstract}
Quadratic programming (QP) solvers are widely used in real-time control and optimization, but their computational cost often limits applicability in time-critical settings. To resolve this, we propose a learning-to-optimize approach using graph neural networks (GNNs) to predict active constraints in the dual active-set solver \texttt{DAQP}. Our method exploits the structural properties of QPs by representing them as bipartite graphs and learns to approximate the optimal active set for effectively warm-starting the solver. Across varying problem sizes, the GNN consistently reduces the number of solver iterations compared to cold-starting, while performance is comparable to a multilayer perceptron baseline. In contrast to the baseline, our GNN-based approach trained on varying problem sizes generalizes to unseen dimensions, demonstrating flexibility and scalability. These results highlight the potential of structure-aware learning to accelerate optimization in real-time applications such as model predictive control.
\end{abstract}

\section{Introduction}
Optimization problems with quadratic objectives and linear constraints, referred to as quadratic programs, form the foundation of numerous applications, including robotics \citep{kuindersma_efficiently_2014}, control \citep{bartlett_quadratic_2002}, and finance \citep{gondzio_parallel_2007, mitra_quadratic_2007}. Given their prevalence, efficient and reliable solution methods are critical for practical applications.

Convex QPs can be solved to global optimality with polynomial complexity \citep{nocedal_numerical_1999}. While interior point methods are highly effective in general, they cannot leverage approximate solutions from related instances, reducing their efficiency for sequential problems.
In contrast, active-set methods naturally support this technique, known as warm starting, by iteratively refining a hypothesis of the active constraint set. Moreover, they are particularly well suited for small- and medium-scale problems \citep{nocedal_numerical_1999}. 

Solving QPs efficiently is critical in for example real-time model predictive control (MPC) \citep{borrelli2017predictive}. These applications provide sequences of closely related optimization problems, making it ideal to exploit solutions from previous iterations. However, naively using the previous solution as a starting iterate can exacerbate worst-case solution time \citep{herceg2015dominant}. Recent works have explored ways of finding warm starts by using machine learning to adapt optimization solvers to problem classes characterized by prior instances \citep{chen2022learning,amos2023tutorial}.

These learning-to-optimize approaches vary in how closely they adhere to conventional solvers. Some focus on tuning hyperparameters \citep{sambharya2024blearning,doerks2025learning}, others learn entire update steps as a black box \citep{andrychowicz_learning_2016}, and hybrid methods combine learned updates with traditional steps to retain convergence guarantees \citep{banert2024accelerated}. \citet{sambharya_end--end_2023} employs an end-to-end approach by predicting initial iterates to warm-start the solver, which are subsequently used to obtain a candidate solution. Prior work has also explored learning penalty parameters for splitting-based QP solvers \citep{ichnowski_accelerating_2021, Saravanos2025}, or predicting search directions for interior-point methods using LSTMs \citep{gao2024ipm}. Further hybrid approaches employ classification trees or k-NN classifiers \citep{klauco_machine_2019}, neural networks \citep{chen_large_2022} or transformers \citep{zinage_transformermpc_2024} to predict active constraints. However, most machine-learning-based approaches for active-set methods ignore the structural properties inherent in optimization problems \citep{ klauco_machine_2019, chen_large_2022, zinage_transformermpc_2024}.

\begin{figure}[t]
    \centering
    \includegraphics[width=1\textwidth]{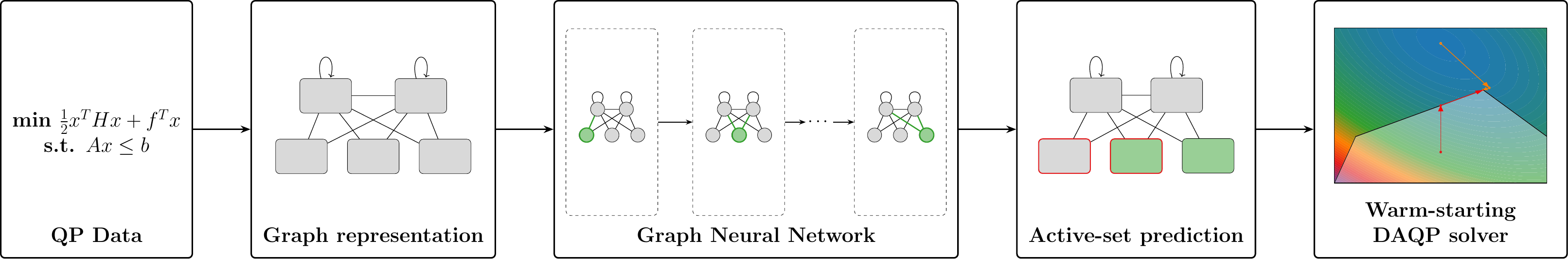}
    \caption{Method overview: The problem is formulated as a QP and represented as a graph, which serves as input to the Graph Neural Network for node-level prediction. The resulting prediction is then used to warm-start the \texttt{DAQP} solver.}
    \label{fig:flowchart}
\end{figure}

In contrast, we investigate how graph neural networks (GNN) accelerate the active-set method for solving QPs. This technique leverages the underlying problem structure using graph representations similar to \citet{sjolund_graph-based_2022,hausner_neural_2024, hausner_learning_2025}. 
By predicting the active set, our proposed approach provides an improved initial guess compared to conventional initialization strategies, enabling the solver to converge in fewer iterations and thereby reducing computation time.
The method overview is depicted in Figure~\ref{fig:flowchart}.
GNNs are particularly well-suited for this task since they inherently capture the structural properties of QPs. They are invariant to node permutations, corresponding to reordering constraints, and can exploit sparsity patterns effectively \citep{cappart_combinatorial_2023}. Previous work has transformed QPs into graph representations to study the existence of GNNs capable of mapping problems to optimal solutions \citep{chen_expressive_2024}, but lacks extensive experimental validation, a gap we address in this paper.

The main contributions of this paper are (i) the first application of GNNs to accelerate QP solving by learning the active set, and (ii) a comprehensive experimental evaluation demonstrating the practical benefits of this approach over multilayer perceptrons, including size generalization.
\section{Background}
In this paper,  we use machine learning to accelerate solving convex quadratic programs (QPs) with linear inequality constraints,
\begin{align} \label{eq:QP}
\begin{split}
    \underset{x \in \mathbb{R}^n}{\text{minimize}} \quad \frac{1}{2}x^THx+f^Tx, \qquad
    \text{subject to } \quad Ax \leq b.
\end{split}
\end{align}
The objective function is defined by the positive definite matrix $H \in \mathbb{R}^{n \times n}$  and the vector $f \in \mathbb{R}^n$, while the feasible region is defined by $A \in \mathbb{R}^{m \times n}$ and $b \in \mathbb{R}^m$. 
Convexity ensures the global optimum can be found efficiently, with computational difficulty comparable to linear programs \citep{nocedal_numerical_1999}.

Many applications require solving a sequence of similar QPs, where only specific parameters change between iterations. We consider problems where the parameter $\eta\in \mathbb{R}^p$ only affects the linear term in the objective function $f$ and the right-hand side of the constraints $b$, i.e.
\begin{align} \label{eq:QP_parametrized}
\begin{split}
    \underset{x \in \mathbb{R}^n}{\text{minimize}} \quad \frac{1}{2}x^THx+f(\eta)^Tx, \qquad 
    \text{subject to } \quad Ax \leq b(\eta).
    \end{split}
\end{align}
Hence, the overall problem structure is fixed across problem instances of equal dimensions.
\subsection{Active-set method}
The active-set method is an iterative solution method for QPs aiming to identify the active set \mbox{$\mathcal{A^*} \subseteq \{1, 2 \dots, m \}$}, consisting of all inequality constraints that hold with equality, at an optimal solution $x^*$. Knowing the active set simplifies the optimization problem by transforming an inequality-constraint problem into one with a reduced set of equality constraints, as stated in Lemma 3.1 in \citet{arnstrom_real-time_2023}. The KKT optimality conditions of this reduced problem form a linear system, which can be solved in a single iteration via matrix factorization techniques.

To identify the active set $\mathcal{A}^*$, a working set $\mathcal{W}$ serves as an approximation of the active set. It is updated iteratively by adding or removing one constraint per iteration until the true active set of an optimal solution $x^*$ is found and the linear system can be solved in a single step.
Solvers initialized with an empty working set $\mathcal{W}_0 = \emptyset$ are cold-started, while warm-started solvers begin with a non-empty working set $\mathcal{W}_0 \neq \emptyset$, which can significantly reduce the number of iterations needed to identify the active set and thereby the optimum \citep{otta_measured-state_2015,arnstrom_real-time_2023}.
While the improvement is most pronounced when the initial working set $\mathcal{W}_0$ closely matches the true active set, even partial overlap can noticeably reduce the iteration count. This motivates a learning-to-optimize approach: a perfect prediction is hard (essentially equivalent to solving the problem), but leveraging prior instances can yield a close approximation and practically significant acceleration.

\subsection{Dual active-set solver}
To benefit from warm-starting, we apply the dual active-set solver proposed by \citet{arnstrom_dual_2022}. Unlike primal solvers, which require a feasible initial iterate $x_0$, dual solvers are easier to warm-start, as dual non-negativity always allows $\lambda_0 = \mathbf{0}$ as a feasible starting point. 
The \texttt{DAQP} solver\footnote{\url{{https://github.com/darnstrom/daqp}}} follows Algorithm~\ref{alg:dual_active_set_method} (Appendix~\ref{ch:algorithm}) from \cite{arnstrom_dual_2022} and leverages an $LDL^T$ factorization for efficient updates. Since the working set changes by at most one constraint per iteration, the lower unit triangular matrix $L$ and diagonal matrix $D$ are updated via rank-one modifications, reducing complexity. Singularity is detected directly from the diagonal of $D$, and the computations rely on forward/backward substitution, avoiding explicit matrix inversion. The reuse of previous computations depends on where in the ordered representation of the working set $\mathcal{W}$ the modification occurs, changes near the end allow most of the factorization to be reused, whereas earlier changes require recomputing a larger portion during the substitution steps.

\subsection{Graph Neural Networks}
\label{ch:GNN}
Graph neural networks (GNN) are neural network architectures operating on graphs. A graph is defined by the vertex set $V$ and the directed edge set $E \subseteq V \times V$. Each vertex $s \in V$ is assigned a feature vector $x_s \in \mathbb{R}^{d_v}$, while each directed edge $e_{ts} \in E$ from vertex $t$ to $s$ is associated with an edge feature vector $z_{ts} \in \mathbb{R}^{d_e}$.
The model consists of multiple GNN layers, each updating vertex and edge features. In this paper, we follow the framework of message-passing GNNs \citep{battaglia_relational_2018}. The update is performed by the permutation-invariant aggregation function $\bigoplus$ acting on the neighborhood of each vertex, and the learnable functions $\phi$ and $\psi$. The graph topology remains fixed throughout the forward pass, while vertex and edge features are updated at each layer.

The update of layer $l$ begins with computing the edges features for the subsequent layer $l+1$ using the parametrized message function $\phi_{\theta_z^{(l)}}$
\begin{align} \label{eq:edge_feature_update}
    z_{ts}^{(l+1)} \coloneq \phi_{\theta_z^{(l)}}\left(z_{ts}^{(l)},x_t^{(l)},x_s^{(l)}\right),
\end{align}
where the output is referred to as message.
In the next step, all incoming messages to a vertex $s$ are aggregated by applying the aggregation function $\bigoplus$ over the neighborhood $\mathcal{N}_s= \{ t \; \vert \;  (t,s) \in E \}$,
\begin{align} \label{eq:aggregation_function}
    m_s^{(l+1)} \coloneq \bigoplus_{t\in \mathcal{N}_s} z_{ts}^{(l+1)}.
\end{align}
Common choices for $\bigoplus$ include summation, maximization, and averaging.
The outcome $m_s^{(l+1)}$ serves as input for the vertex feature update for layer $l+1$ by the parametrized function $\psi_{\theta_x^{(l)}}$
\begin{align} \label{eq:vertex_feature_update}
    x_s^{(l+1)} \coloneq \psi_{\theta_{x}^{(l)}}\left(x_s^{(l)},m_s^{(l+1)}\right).
\end{align}
The update functions $\phi$ and $\psi$ are parametrized by learnable parameters $\theta$ \citep{bronstein_geometric_2021}.
To apply this method to solve QPs, we next describe how they are represented as graphs.

\section{Method} \label{ch:methods}
The proposed GNN-based approach for solving QPs comprises three key components: the graph representation of the data, the learned mapping, and the final model architecture.

\subsection{Graph representation} \label{ch:graph_representation}
We represent the underlying QPs as bipartite graphs following \citet{chen_expressive_2024}.

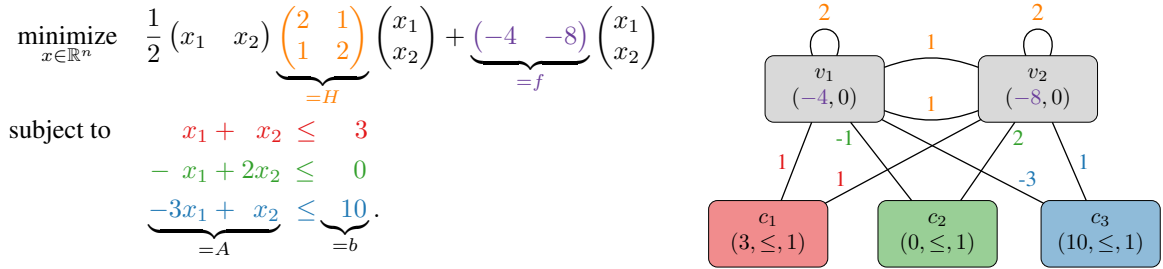
\begin{figure}[t]
\begin{minipage}{0.6\textwidth}
    \small
    $
    \begin{aligned}
        \underset{x \in \mathbb{R}^n}{\text{minimize}} &\quad 
        \frac{1}{2} \begin{pmatrix}x_1 & x_2\end{pmatrix}
        \underbrace{\textcolor{darkorange}{\begin{pmatrix}2 & 1 \\ 1 & 2\end{pmatrix}}}_{\textcolor{darkorange}{=H}}
        \begin{pmatrix}x_1 \\ x_2\end{pmatrix}
        + 
        \underbrace{\textcolor{darkpurple}{\begin{pmatrix}-4 & -8\end{pmatrix}}}_{\textcolor{darkpurple}{=f}}
        \begin{pmatrix}x_1 \\ x_2\end{pmatrix} \\
        \text{subject to } & \quad \quad \: \:\textcolor{darkred}{x_1+ \: \, \, x_2 \: \leq \quad \, 3}\\
        & \quad \textcolor{darkgreen}{ \:-\:\: x_1 + 2x_2 \: \leq \quad \,0} \\
        & \quad \underbrace{\textcolor{darkblue}{-3x_1+ \: \:x_2}}_{=A} \: \:\textcolor{darkblue}{\leq} \underbrace{\: \:\:\textcolor{darkblue}{10}}_{=b}.
    \end{aligned}$
\end{minipage} 
  \begin{minipage}{.4\textwidth}
    \centering
    \resizebox{\linewidth}{!}{
    \begin{tikzpicture}[
        node distance=2cm, 
        red/.style={fill=darkred!50,text width=1.5cm,rectangle, draw=black, rounded corners, font=\sffamily\small, minimum height=1cm, minimum width = 1.75cm, text centered,inner sep=2mm, outer sep=0pt},
        green/.style={fill=darkgreen!50,text width=1.5cm,rectangle, draw=black, rounded corners, font=\sffamily\small, minimum height=1cm, minimum width = 1.75cm, text centered,inner sep=2mm, outer sep=0pt},
        blue/.style={fill=darkblue!50,text width=1.5cm,rectangle, draw=black, rounded corners, font=\sffamily\small, minimum height=1cm, minimum width = 1.75cm, text centered,inner sep=2mm, outer sep=0pt},
        gray/.style={fill=gray!30,text width=1.5cm,rectangle, draw=black, rounded corners, font=\sffamily\small, minimum height=1cm, minimum width = 1.75cm, text centered,inner sep=2mm, outer sep=0pt},
        edge/.style={draw, semithick, black, -},
        edge_labels/.style={font=\small, midway},
        every loop/.style={in = 110, out = 70,looseness = 4}
    ]

    % Nodes
    \node[gray] (0) {$v_1$ \par $(\textcolor{darkpurple}{-4},0)$};
    \node[gray, right = 1.5cm of 0] (1) {$v_2$ \par $(\textcolor{darkpurple}{-8},0)$};
    \node[red, below left= 1.25cm and -1cm of 0] (2) {$c_1$ \par $(3,\leq,1)$};
    \node[green,below right = 1.25cm and -0.1cm of 0] (3) {$c_2$ \par $(0,\leq,1)$};
    \node[blue,below right = 1.25cm and 2.5cm of 0] (4) {$c_3$ \par $(10,\leq,1)$};

    % Edges
    \path[edge] (0) edge[loop above] node[above] {\textcolor{darkorange}{2}} (0);
    \path[edge] (1) edge[loop above] node[above] {\textcolor{darkorange}{2}} (1);
    \path[edge] (0) to [bend right=20] node[edge_labels, above] {\textcolor{darkorange}{1}} (1);
    \path[edge] (1)  to [bend right=20] node[edge_labels, above] {\textcolor{darkorange}{1}} (0);
    \path[edge] (0) -- (2) node[edge_labels,pos = 0.5, left] {\textcolor{darkred}{1}};
    \path[edge] (0) -- (3) node[edge_labels,pos = 0.2, left] {\textcolor{darkgreen}{-1}};
    \path[edge] (1) -- (3) node[edge_labels,pos = 0.2, right] {\textcolor{darkgreen}{2}};
    \path[edge] (1) -- (4) node[edge_labels, pos = 0.5,right ] {\textcolor{darkblue}{1}};
    \path[edge] ($ (1.south west) + (0.045,0.045) $) -- ($ (2.north east) + (-0.045,-0.045) $) node[edge_labels,pos=0.9, above] {\textcolor{darkred}{1}};
    \path[edge] ($ (0.south east) + (-0.045,0.045) $) -- ($ (4.north west) + (0.045,-0.045) $) node[edge_labels,pos=0.9, above] {\textcolor{darkblue}{-3}};

    \end{tikzpicture}}
  \end{minipage}
\caption{A QP can be represented as a bipartite graph having a set of variable nodes (gray) and a set of constraint nodes (colored), with edge features representing relationships between variables.}
\label{fig:graph_representation_example}
\end{figure}

\begin{definition} \label{def:graph_representation}
    The graph representation of a QP is defined by $G=(W\cup C,\:E_W \cup E_C)$, where
    \begin{itemize}
        \item the set $W=\{1, \dots, n\}$ represents the variable nodes. The $i$-th vertex represents the $i$-th variable of the QP and the assigned feature vector $x_s \in \mathbb{R}^{w}$ captures the corresponding component of $f$ as well as the node type.
        \item the set $C = \{n+1, \dots, m\}$ represents the constraint nodes. The $j$-th vertex represents the $j$-th constraint of the QP and the assigned feature vector $x_s \in \mathbb{R}^{c}$ captures the corresponding component of $b$ as well as the direction of the inequality sign and the node type.
        \item the edges contained in $E_W$ and their feature vectors are given by the adjacency matrix $H$, which defines the quadratic term in the QP.
        \item the edges contained in $E_C$ and their feature vectors are given by the adjacency matrix $A$ and show the relations between the variable and constraint nodes.
    \end{itemize}
\end{definition}
This is exemplified in Figure~\ref{fig:graph_representation_example}.

 \subsection{Learning problem}
The key idea is to learn the parameters $\theta$ of a mapping $p_\theta$ from a problem instance represented by a graph $G$ to the resulting active set $\mathcal{A}^*$
\begin{equation}
    p_\theta:\mathcal{G} \rightarrow \mathcal{P}(\mathbb{N}_m), \quad p_\theta(G) \approx \mathcal{A}^*,
\end{equation} 
where $\mathcal{G}$ denotes the set of all graphs with $m$ variable nodes and $\mathcal{P}$ the power set. This corresponds to a node classification task on the graph $G$, determining whether each constraint vertex is active or not. Finding this mapping is not trivial, since the optimal solution $x^*$ must be known to determine the active set $\mathcal{A}^*$ \citep{zinage_transformermpc_2024}. 
The model is trained in a supervised fashion, where ground-truth active sets are obtained by solving problem instances to optimality using the cold-started \texttt{DAQP} solver, and the resulting predictions are used to warm-start the solver at inference time.

This approach has the potential to automate and improve on the non-trivial task of finding a good initial guess, while retaining the theoretical convergence guarantees of conventional solvers. Moreover, the graph representation implies permutation equivariance, such that the order of constraints does not affect the model.
Finally, GNNs can operate on graphs of varying sizes, making them easily adaptable to different problem structures~\citep{chen_expressive_2024}.

\subsection{Model architecture} \label{ch:model_architecture}
The only restriction our approach places on the GNN architecture is to operate on the bipartite graph representation in Definition~\ref{def:graph_representation}. As a proof of concept, this paper uses Local Extrema Convolution (LEConv) layers \citep{ranjan_asap_2020}, leaving the design of tailored layers to future work.

In the LEConv architecture, the edge feature update for layer $l+1$ is defined by the function $\phi_{\theta_z^{(l)}}$ in Equation~\eqref{eq:edge_feature_update} as
\begin{align}
    z_{ts}^{(l+1)} \coloneq z_{ts}^{(l)} \cdot \left(x_s^{(l)} W_2 - x_t^{(l)} W_3 \right),
\end{align}
where $z_{ts}^{(l)}$ represents the feature of the directed edge from node $t$ to $s$ in layer $l$, $x_s^{(l)}$ and $x_t^{(l)}$ are the corresponding vertex features, and $W_j$ are learnable weight matrices.
Sum aggregation as in Equation~\eqref{eq:aggregation_function} yields the message for node $s$ in layer $l+1$
\begin{align}
    m_s^{(l+1)} \coloneq \sum_{t \in \mathcal{N}_s} z_{ts}^{(l)} \left(x_s^{(l)} W_2 - x_t^{(l)} W_3 \right).
\end{align}
The vertex update function $\psi(\cdot)$, as introduced in Equation~\eqref{eq:vertex_feature_update}, then combines the node’s previous feature with the aggregated message using an activation function $\sigma(\cdot)$:
\begin{align}
\begin{split}
    x_s^{(l+1)} & \coloneq \sigma \left( x_s^{(l)} W_1 + m_s^{(l+1)} \right) 
                 = \sigma \left( x_s^{(l)} W_1 + \sum_{t \in \mathcal{N}_s} z_{ts}^{(l)} \left(x_s^{(l)} W_2 - x_t^{(l)} W_3 \right) \right).
\end{split}
\end{align}
The final model consists of three layers, an input layer, a hidden layer of width $128$ and an output layer. All layers share the same architectural structure, presented above. LeakyRELU was chosen as the activation function with a negative slope of $0.1$. A sigmoid activation function is applied to the output layer to interpret the outputs as probabilities of a node being active or not.

The model is trained using the AdamW optimizer with default learning rates and a weighted binary cross-entropy loss, where class weights reflect the empirical class distribution. Early stopping based on validation loss is applied, terminating training if the loss does not improve by at least $0.001$ over five consecutive epochs, with the best-performing parameters retained. The threshold for predicting active constraint is set by performing a grid search over $[0,1]$ with a step size of $0.1$.

\section{Experiments}
The experiments evaluate the model on synthetic data to assess its ability to predict the active set and compare it against a MLP. Subsequently, the GNN is applied to datasets with varying problem sizes and an inverted pendulum control problem. The results examine the predictive accuracy and the impact on solver efficiency. All experiments are conducted on a single NVIDIA Titan Xp GPU.
\subsection{Synthetic data generation}  \label{ch:problem_generation_synthetic}
The first experiments use synthetic data parameterized by $\eta \, \in \mathbb{R}^p $ as in Equation~\eqref{eq:QP_parametrized}. For each problem dimension, the matrices $H$ and $A$ are generated once and kept constant across all instances, while only the parameter $\eta$ varies. To ensure that the matrix in the objective function $H$ is symmetric and positive definite, it is generated using the matrix $M \in \mathbb{R}^{n \times n}$ with \mbox{$H = MM^T$}. If required, sparsity patterns can be applied to the matrices. The right-hand side of the constraints and the linear term in the objective function are defined by affine transformations $f : \mathbb{R}^{p} \rightarrow \mathbb{R}^n$ and $b : \mathbb{R}^{p} \rightarrow \mathbb{R}^m$, 
\begin{align}
   f(\eta) = \hat{f}+F\eta,  \quad \quad \quad  b(\eta) = \hat{b} -AT\eta,
\end{align}
where $\hat{f} \in  \mathbb{R}^n$, $F \in \mathbb{R}^{n \times p}$, $\hat{b} \in  \mathbb{R}^m$ and the transformation matrix $T \in \mathbb{R}^{n \times p}$ ensures primal feasibility of $x = T \eta$.
All variables are sampled from a standard normal distribution $\mathcal{N}(0,1)$, with the exception of $\hat{b}$, which is drawn from a uniform distribution over $[0,1)$ to ensure that the origin is feasible. This guarantees that the \texttt{DAQP} solver used in the implementation finds a feasible solution.

\subsection{Active-set prediction using GNN} \label{ch:exp_active_set_prediction}
To demonstrate the ability of capturing the problem structure using GNNs, an experimental pipeline was implemented following the methods in Section~\ref{ch:methods}. The synthetic data, generated as described in Section~\ref{ch:problem_generation_synthetic}, was converted into graph representations according to Section~\ref{ch:graph_representation} and used as input to the model detailed in Section~\ref{ch:model_architecture}.

\begin{wraptable}{r}{5cm}
\caption{Metrics of the GNN model trained on synthetic data with $n = 10$ (variables) and $m = 40$ (constraints).}
\centering
\begin{tabular}{l c}
\toprule
Metric  & Test \\
\midrule
Accuracy (\%) & 95.43 $\pm$ 0.4\\ 
Precision (\%)& 89.53 $\pm$ 1.4\\
Recall (\%) & 89.74 $\pm$ 1.7\\
F1-score (\%) & 89.62 $\pm$ 0.9\\
\bottomrule
\end{tabular}
\label{table:test_metrics_10v_40c_fixedHA}
\end{wraptable}
The evaluation uses standard classification metrics computed only over constraint nodes, as these are the only relevant nodes in this node-level classification task.
All metrics are reported as percentages and rounded to two decimal places. As a baseline for comparison, a na\"ive model that classifies all nodes as inactive, corresponding to cold-starting the solver, is given.

The dataset contains 5000 instances with $n = 10$ (variables) and $m = 40$ (constraints). Table~\ref{table:test_metrics_10v_40c_fixedHA} shows averages over five runs with different random seeds. The GNN achieves over $89\%$ across all metrics on the test set, clearly outperforming the na\"ive baseline (Accuracy $82.96\%)$, indicating its strong predictive capability. The accuracy is the highest metric due to the unbalanced dataset, as each graph contains more inactive than active nodes.
\subsection{Impact of problem structure on predictive performance}

To assess whether incorporating problem structure through a graph neural network improves predictive performance, we compare it to a standard multilayer perceptron (MLP). For the MLP, all QP components are concatenated into a single input vector by flattening $H \in \mathbb{R}^{n \times n}$ and $A \in \mathbb{R}^{m \times n}$, and appending $f \in \mathbb{R}^n$ and $b \in \mathbb{R}^m$, yielding a vector of size $n^2 + mn + n + m$. The MLP produces binary predictions for all $n+m$ elements, enabling direct comparison with the GNN, though only the $m$ constraint predictions are of primary interest. For each problem size, $2000$ QPs are generated with shared matrices $H$ and $A$ and sparsity is introduced via banded matrices $M$ and $A$. For different problem sizes, new matrices are generated.
It is important to note that an MLP has a fixed input size, such that a separate model needs to be trained for each problem size. 

\begin{wrapfigure}{r}{0.4\textwidth}
    \centering
    \vspace{3mm}
    \includegraphics[width=\linewidth]{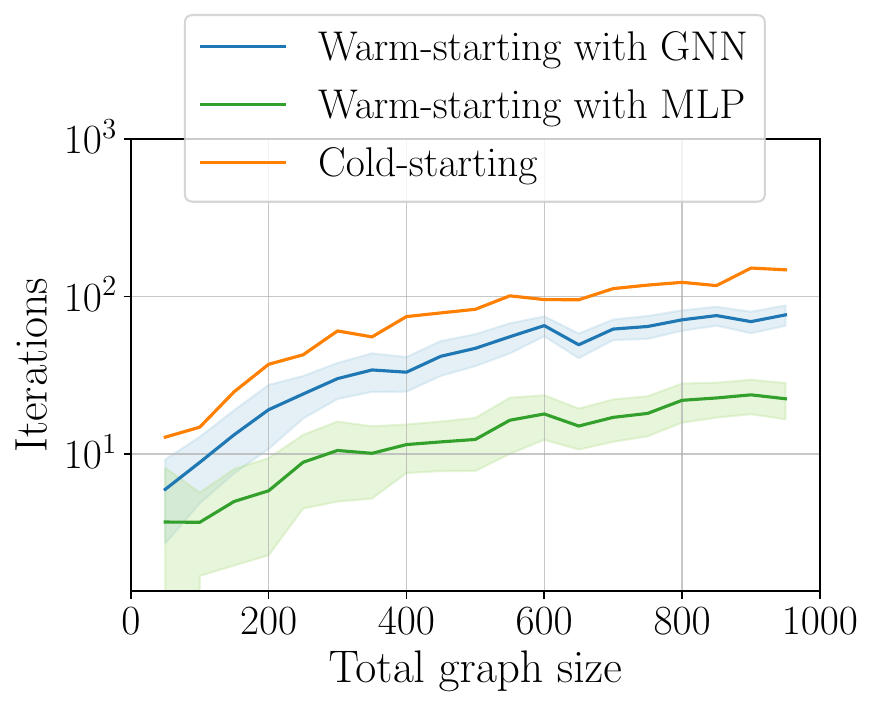}
    \caption{Comparison of iterations when warm-starting the \texttt{DAQP} solver with predictions from our graph neural network, a standard multilayer perceptron and cold-starting the solver without any learned prediction.}
    \label{fig:scaling_plot_iterations}
\end{wrapfigure} 

\begin{figure}[t]
\centering
\includegraphics[width=0.75\textwidth]{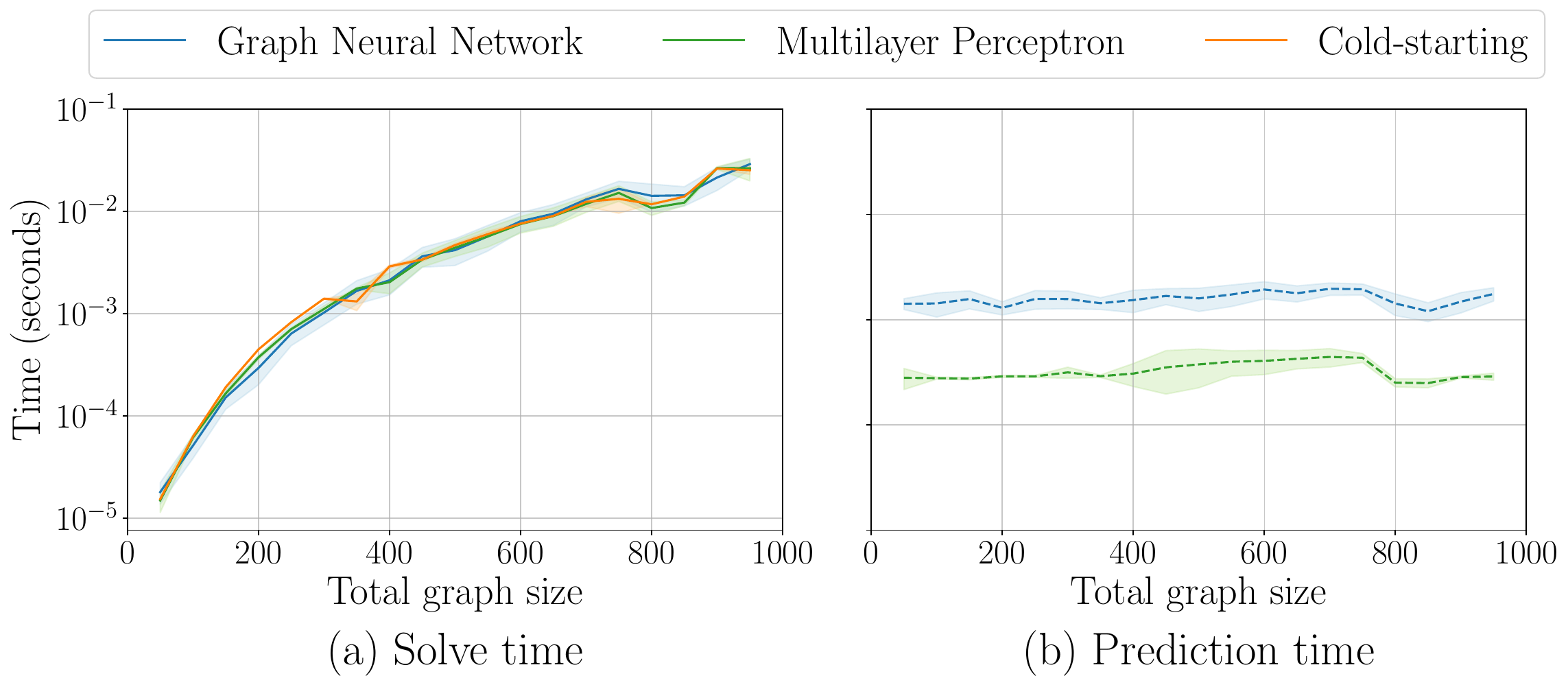}
\caption{Comparison of (a) solve time and (b) prediction time for our graph neural network (blue), a standard multilayer perceptron (green), and the cold-started active set method (orange).}
\label{fig:scaling_plot_time}
\end{figure}

The comparison considers the number of iterations (Figure~\ref{fig:scaling_plot_iterations}) and total solve time (Figure~\ref{fig:scaling_plot_time}) for increasing problem sizes, while maintaining a fixed $1{:}4$ ratio between variable and constraint nodes. All values represent the mean over five independent runs per problem size.

Figure~\ref{fig:scaling_plot_iterations} shows that warm-starting the \texttt{DAQP} solver with predictions from either model substantially reduces the iteration count compared to cold-starting. The iteration growth remains sublinear across all approaches.

Figure~\ref{fig:scaling_plot_time} compares solve and prediction times across varying problem sizes.
The solve time is measured using the built-in timing of the \texttt{DAQP} solver, combining initialization and computing time, ensuring consistent evaluation conditions. When a model predicts an active set, it is provided as input to the solver. For cold-starting, no such initialization is available, and the problem is solved without prior knowledge. The prediction time refers to the forward pass during testing, and all results are averaged over five independent runs.
The solve times for all models scale similarly, approximately proportional to $x^{3/2}$ as shown in Appendix~\ref{ch:log_log}, with the GNN consistently reducing solve time compared to cold-starting, as further illustrated in the next section.
Both models exhibit approximately constant prediction times as the problem size increases, with the MLP achieving predictions nearly an order of magnitude faster than the GNN due to its simpler architecture. This simplicity, however, comes at the cost of generalization. The weight-sharing and structure-aware design of the GNN is precisely what enables it to generalize across problem sizes, as shown in the next section.

\subsection{Generalization across variable problem sizes}
\begin{figure}[b]
\centering
 \includegraphics[width=0.75\textwidth]{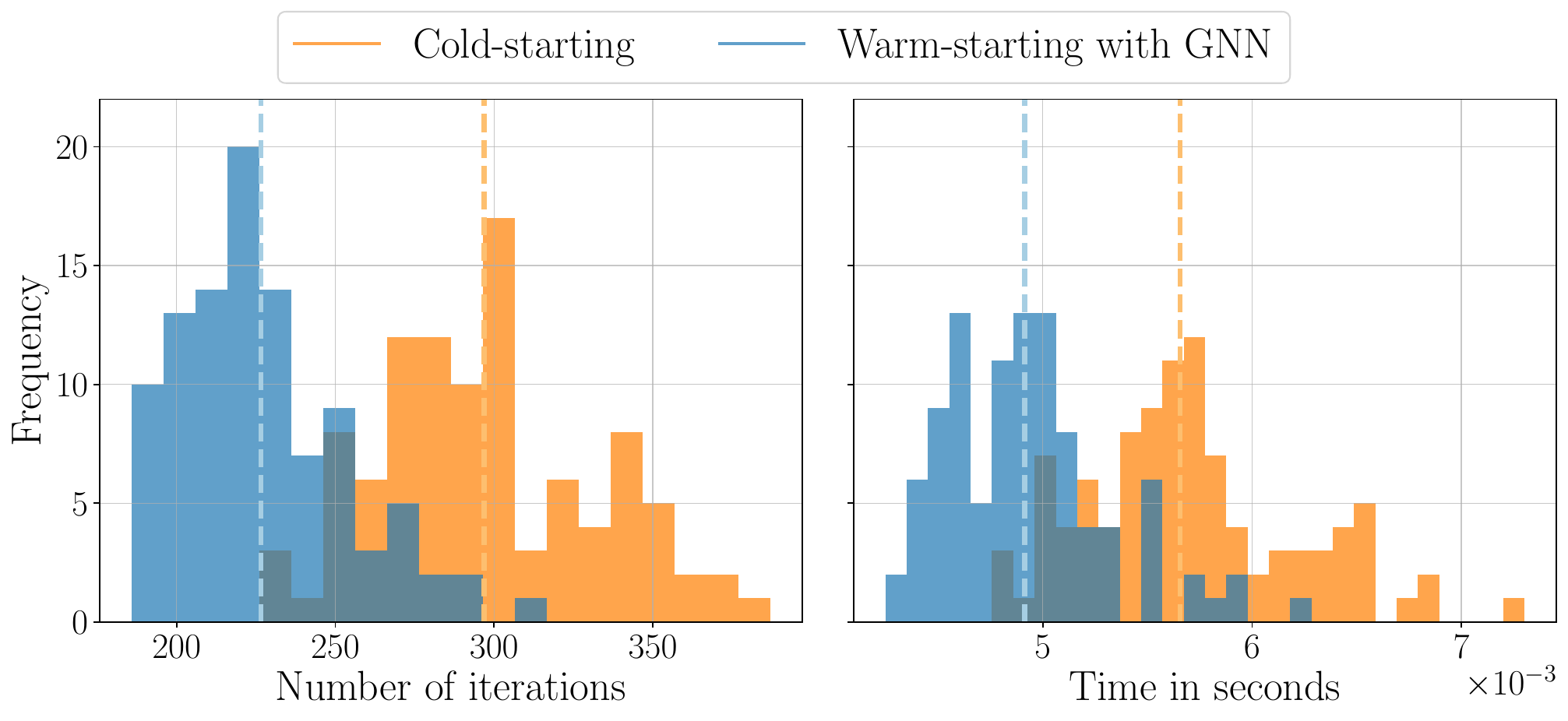}
\caption{Comparison of iterations (left) and solve time (right) of cold-starting (orange) and warm-starting the \texttt{DAQP} solver using the GNN (blue) on problem instances with $100$ variables and $200$ constraints. The dashed line represents the mean.}
\label{fig:it_time_multi}
\vspace{-0.6cm}
\end{figure}

Although the solve time for the GNN and MLP is comparable, and the MLP achieves a slightly lower iteration count, its fixed input size limits its applicability to varying problem sizes. In contrast, GNNs naturally handle graphs of arbitrary size, making them well-suited for scenarios where the number of variables and constraints varies across instances. This flexibility enables generalization across heterogeneous problem structures, offering clear practical advantages.

To illustrate this, the GNN was trained on graphs of size $[60,120,180]$ and tested on graphs of size $300$, with a single pair of matrices $H$ and $A$ shared across all problems of the same dimension. As Figure~\ref{fig:it_time_multi} shows, the model still achieves substantial reductions in iterations and solve time, demonstrating that models trained on smaller problem sizes can generalize effectively to larger ones, reducing the need for retraining and providing practical efficiency gains.
\subsection{Effect of predictive model on solver efficiency}
To evaluate the GNN's impact on the solver performance under realistic conditions, we apply our method to the control problem of an inverted pendulum on a moving cart,  generated using the \texttt{lmpc} package\footnote{\url{https://github.com/darnstrom/lmpc}}. The package produces model predictive control (MPC) data for linear systems, resulting in problems of the form
\begin{align} \label{eq:pendulum}
\begin{split}
    \underset{u_0,  \dots, u_{N_c-1}}{\text{minimize}} &\quad \frac{1}{2} \sum_{k=0}^{N_p-1} ((C z_k-r)^T Q (Cz_k-r) + u_k^T R u_k + \Delta u_k^T R_r \Delta u_k) \\
    \text{subject to } &\quad z_{k+1} = Fz_k + Gu_k, \qquad \quad \; k = 0,\dots ,N_p-1 \\
    & \quad z_0 = \hat{z} \\
    & \quad \underline{b} \leq A_z z_k + A_u u_k \leq \overline{b}, \qquad k = 0 , \dots ,N_p-1,
\end{split}
\end{align}
where $z_k$ and $u_k$ are the system state and control action, respectively, at time step $k$. The current state $\hat{z}$ and a set point $r$ make this a parametric optimization problem. This problem can be condensed (see Chapter 2 in \citep{arnstrom_real-time_2023}) to yield QPs of the form \eqref{eq:QP}. Here, the decision variable $x$ contains the control actions $u_0,\dots,u_{N_c-1}$, and the linear term $f$ in the objective function and the right-hand side $b$ of the constraints are affine functions of $\hat{z}$ and $r$, while $H$ and $A$ remain fixed.

Two datasets with different control horizons are compared, both use a prediction horizon of $N_p = 50$. The first dataset has a control horizon of $N_c = 5$, resulting in $206$ constraints, while the second uses $N_c = 50$ with $m = 296$ constraints. Note that the higher number of constraints in the MPC setting, reduces the fraction of active constraints by an order of magnitude compared to the synthetic data. The GNN is trained on the generated data, results are averaged over five independent runs, and a regularization term is added to encourage sparsity in the predicted active sets.

As in previous experiments, we compare our GNN-based warm-starting against a cold-started solver. While conventional warm-starting, 
is effective when consecutive parameters are similar, this does not always hold in MPC. In reference tracking, for instance, setpoint changes can vastly change parameters relative to the previous problem, causing conventional warm-starting to even perform worse than cold-starting \citep{herceg2015dominant}. Hence, our method complements the conventional warm-starting MPC solvers, yielding speedups even in scenarios where the latter fails. 

\begin{figure}[t]
\centering
\includegraphics[width=0.75\textwidth]{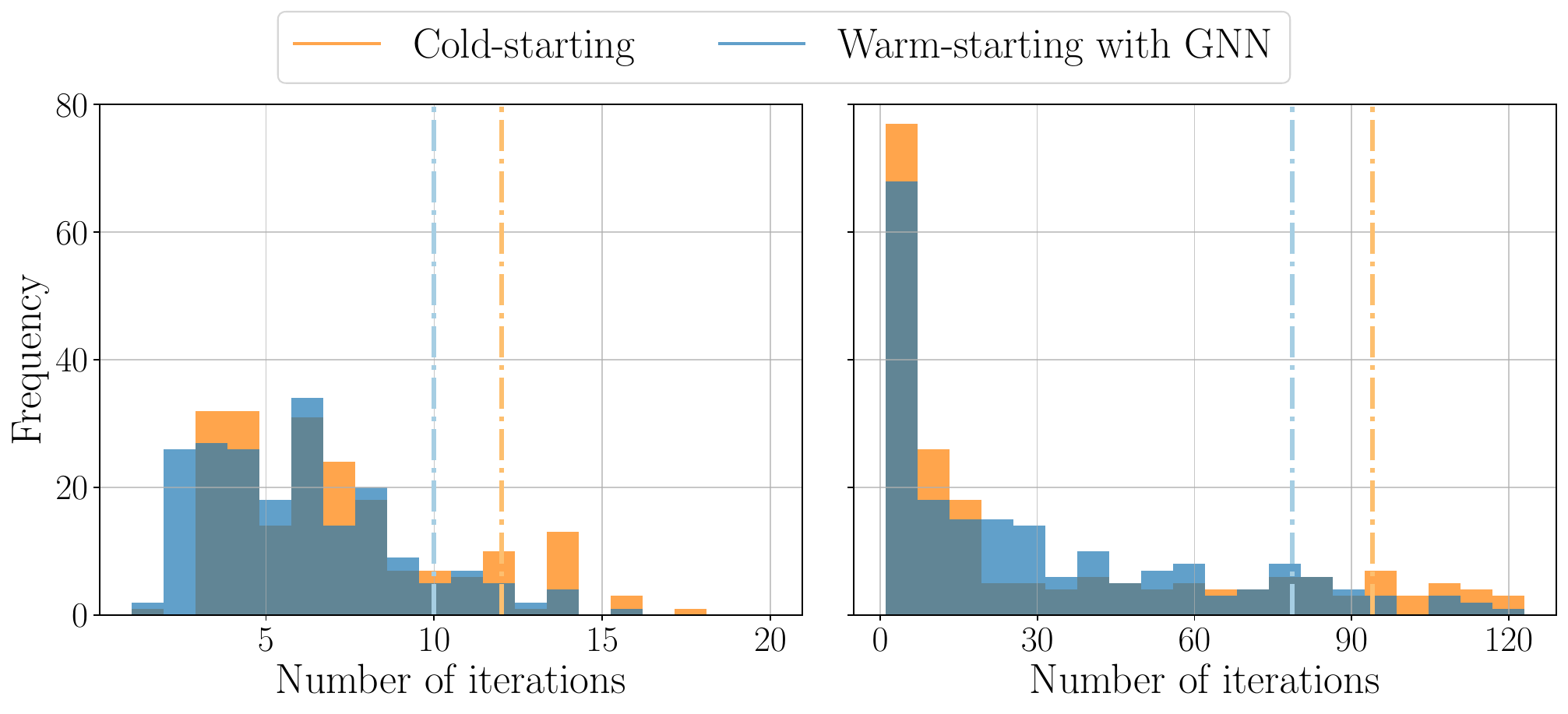}
\caption{Comparison of iterations 
of cold-starting (orange) and warm-starting the \texttt{DAQP} solver using the GNN (blue) on problem instances with control horizon $N_c =5$ (left) and $N_c = 50$ (right). The prediction horizon is $N_p = 50$ in both cases and the constraints are $206$ and $296$ respectively. The dash-dotted line refers to the $90\%$ percentile.}
\label{fig:it_time_lmpc}
\end{figure}
Specifically for the inverted pendulum, the control $u$ is a force applied to the cart,
\begin{wraptable}{r}{9.3cm}
\centering
\caption{Iteration reduction on dataset with $N_c = 50$.}
\begin{tabular}{l c c c c c c c}
\toprule
Percentiles & 10\% &25\% & 50 \% & 75\% & 90\% \\
\midrule
Iterations cold-start & 3 & 4 & 13 & 55.5 & 94\\ 
Iterations warm-start & 3 & 4 & 19 & 47.6 & 78.7\\
\bottomrule
\end{tabular}
\label{table:iteration_reduction_lmpc}
\vspace{-0.3cm}
\end{wraptable}
and the state is $z=(p,\dot{p},\phi,\dot{\phi})$, where $p$ is the cart position and $\phi$ is the pendulum angle. We have the actuation constraint $|u|\leq 1$ and the state constraints $|p|\leq 10$ and $\left|\phi\right| \leq \frac{\pi}{4}$. The specific input parameters used in this implementation can be found in the repository\footnote{\url{https://github.com/ellaschmidtobreick/acc_daqp}}.

Figure~\ref{fig:it_time_lmpc} shows the change in solver iterations.

Warm-starting the QP solver with the active set predicted by the GNN reduces the number of iterations, with an average reduction of $1.0$ iterations on the first dataset and $1.6$ iterations on the second dataset. The dotted line in the figure indicates the $90\%$ percentile, showing greater reduction for problems requiring more iterations when cold-started.

This trend is further confirmed in Table~\ref{table:iteration_reduction_lmpc}, which breaks down the iteration reduction across different percentile thresholds for the data set with control horizon $N_c = 50$: the harder the problem for the cold-started solver, the greater the reduction in number of iterations, while problems that converge quickly see minimal benefit. These iteration savings translate directly into reduced solve times (Appendix~\ref{MPC_time_figure}), consistent with the findings of earlier experiments.

\section{Conclusion}
This work presented a learning-to-optimize approach using GNNs to predict the active set in dual active-set solvers for quadratic programs. By representing QPs as bipartite graphs, the model captures the problems' structural properties and can learn effective initial active sets for warm-starting. The experiments demonstrate consistent reductions in solver iterations and solve time, with the largest gains for problems requiring many iterations when cold-started.
In contrast to the MLP baseline, our approach naturally handles and scales well with increasing problem size. The GNN exhibits strong size generalization, effectively transferring to larger graphs even when trained only on smaller instances, highlighting its flexibility and practical applicability.

These results indicate that incorporating learned structure-aware predictions can significantly accelerate optimization in sequential or real-time settings. Future work could extend this to other problem classes and further investigate model generalization from smaller to larger problem instances to reduce training cost.
Furthermore, a multi-step approach, first predicting the number of active constraints and then identifying which are active, could improve prediction accuracy and enable faster and more adaptive control in real-world systems.

\acks{This work was funded by the Swedish Research Council, grant number 2024-04130, and supported by the Wallenberg AI, Autonomous Systems and Software Program (WASP) funded by the Knut and Alice Wallenberg Foundation. The computations were enabled by resources provided by the National Academic Infrastructure for Supercomputing in Sweden (NAISS), partially funded by the Swedish Research Council through grant agreement no. 2022-06725.}

\bibliography{refs}

\newpage
\appendix
\section{Algorithm of the \texttt{DAQP} solver} \label{ch:algorithm}
\input{algorithm}
\newpage
\section{Scaling results on log-log axes} \label{ch:log_log}
\begin{figure}[h]
\centering
\includegraphics[width=0.75\textwidth]{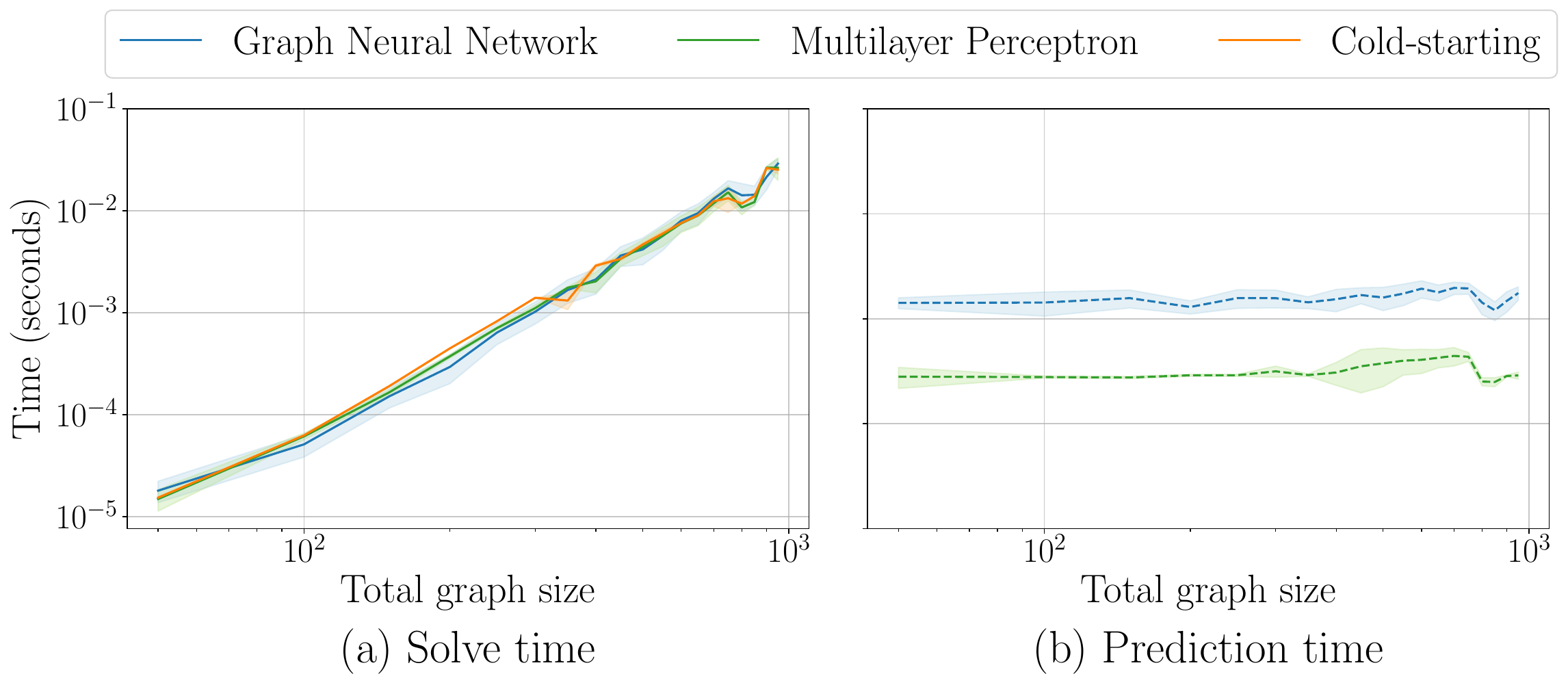}
\caption{Comparison of solve time (a) and prediction time (b) for our graph neural network (blue), a standard multilayer perceptron (green), and the cold-started active set method (orange), shown on logarithmic axes. For the largest graph sizes, the solve time with GNN warm-starting is about five times shorter than the other two, with a negligible prediction time.}
\label{fig:scaling_plot_time_log_log}
\end{figure}

\begin{figure}[h]
    \centering
    \includegraphics[width=0.4\textwidth]{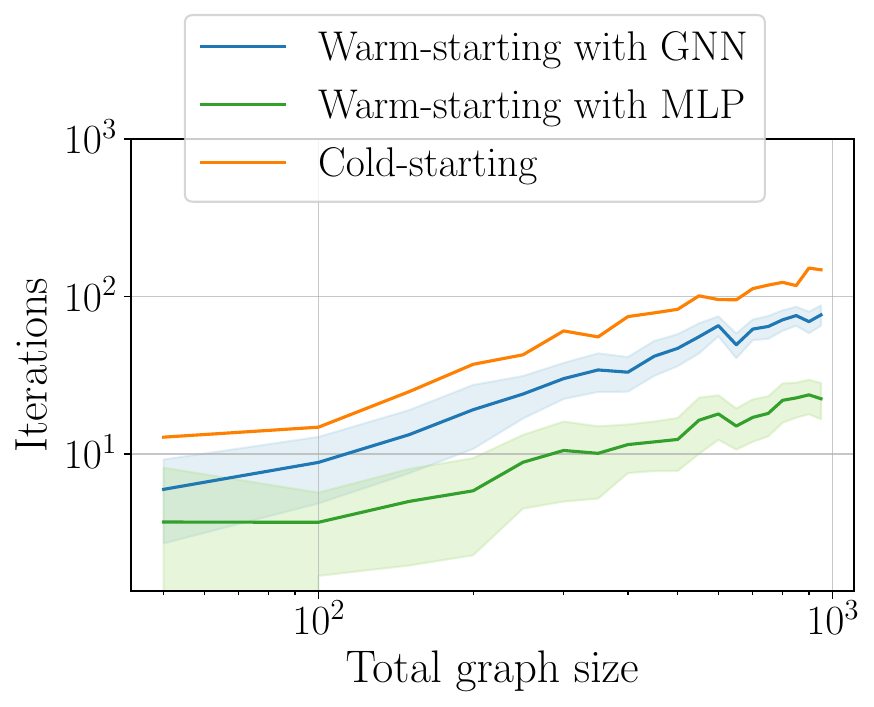}
    \caption{Comparison of iterations when warm-starting the \texttt{DAQP} solver with predictions from our graph neural network, a standard multilayer perceptron and cold-starting the solver without any learned prediction, shown on logarithmic axes.}
    \label{fig:scaling_plot_iterations_log_log}
\end{figure}

\newpage
\section{MPC solve time results} \label{MPC_time_figure}
\begin{figure}[h]
\centering
\includegraphics[width=0.75\textwidth]{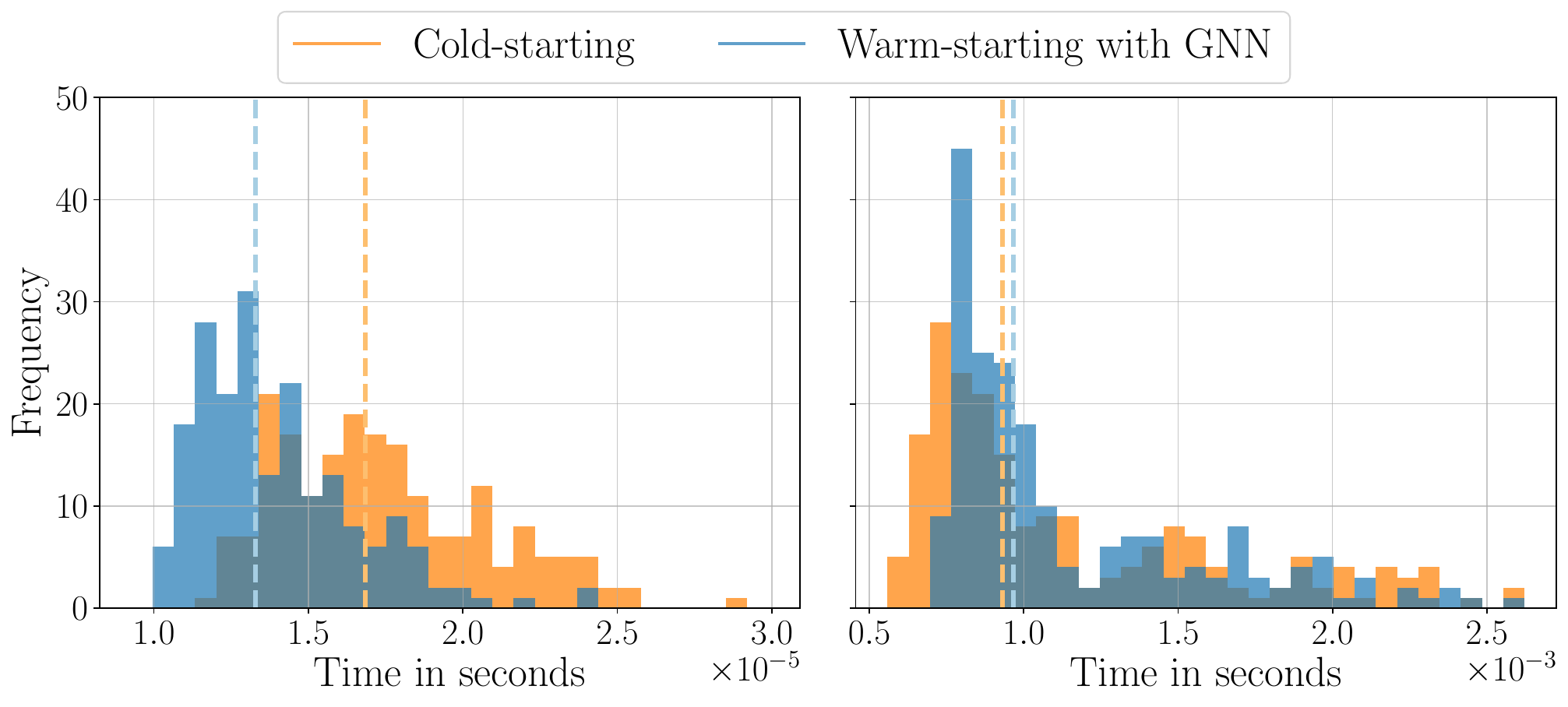}
\caption{Comparison of solve time of cold-starting (orange) and warm-starting the \texttt{DAQP} solver using the GNN (blue) on problem instances with control horizon $5$ (left) and $50$ (right). The prediction horizon is $50$ in both cases and the constraints are $206$ and $296$ respectively. The dash-dotted line refers to the $90\%$ percentile.}
\label{fig:time_lmpc}
\end{figure}

\end{document}

%% file: algorithm.tex
\begin{algorithm}[h]
\caption{Dual active-set method for solving a dual QP \citep{arnstrom_dual_2022}}
\label{alg:dual_active_set_method}
\begin{algorithmic}[1]
\REQUIRE $M,d,v,R^{-1}, \mathcal{W}_0,\lambda_0$
\ENSURE $x^*,\lambda^*,\mathcal{A}^*$
\WHILE{true}
    \IF{$M_{\mathcal{W}}M_{\mathcal{W}}^T$ is nonsingular}
        \STATE Solve $M_{\mathcal{W}}M_{\mathcal{W}}^T\lambda^*_{\mathcal{W}} = - d_{\mathcal{W}}$
        \IF{$\lambda^* \ge 0$} 
            \STATE $\mu_{\overline{\mathcal{W}}} \gets M_{\overline{\mathcal{W}}}M_{\mathcal{W}}^T\lambda^*_{\mathcal{W}} + d_{\overline{\mathcal{W}}}$ 
            \STATE $\lambda \gets \lambda^*$ \algcomment{$\lambda^*$ dual feasible}
            \IF{$\mu \ge 0$}
                \STATE \textbf{break} \algcomment{$x^*$ primal feasible}
            \ELSE 
                \STATE $j \gets \arg\min_{i \in \overline{\mathcal{W}}} [\mu]_i$ \algcomment{$x^*$ not primal feasible}
                \STATE $\mathcal{W} \gets \mathcal{W} \cup \{j\}$ 
            \ENDIF
        \ELSE 
            \STATE $p \gets \lambda^* - \lambda$ \algcomment{$\lambda^*$ not dual feasible}
            \STATE $\mathcal{B} \gets \{i \in \mathcal{W} \mid [\lambda^*]_i < 0\}$
            \STATE $(\lambda, \mathcal{W}) \gets \textsc{FixComponent}(\lambda, \mathcal{W}, \mathcal{B}, p)$
        \ENDIF
    \ELSE
        \STATE Solve $M_{\mathcal{W}}M_{\mathcal{W}}^T p_{\mathcal{W}} = 0$
        \STATE $\mathcal{B} \gets \{i \in \mathcal{W} \mid [p]_i < 0\}$
        \STATE $(\lambda, \mathcal{W}) \gets \textsc{FixComponent}(\lambda, \mathcal{W}, \mathcal{B}, p)$
    \ENDIF
\ENDWHILE
\STATE $x^* \gets -R^{-1}(M_{\mathcal{W}}^T \lambda^*_{\mathcal{W}} + v)$
\RETURN $(x^*,\lambda^*,\mathcal{W})$
\end{algorithmic}
\end{algorithm}
\vspace{-0.8cm}
\begin{algorithm}
\begin{algorithmic}[1]
\setcounter{ALC@line}{25}
\vspace{0.1cm}
    \STATE{\textbf{procedure }\textsc{FixComponent}}{($\lambda, \mathcal{W}, \mathcal{B}, p$)}
    \STATE \hspace{1em} $j \gets \arg\min_{i \in \mathcal{B}} \left(-\frac{[\lambda]_i}{[p]_i}\right)$
    \STATE \hspace{1em}$\mathcal{W} \gets \mathcal{W} \setminus \{j\}$
    \STATE \hspace{1em}$\lambda \gets \lambda - \frac{[\lambda]_j}{[p]_j} p$
    \RETURN $(\lambda, \mathcal{W})$
%\ENDFUNCTION
\end{algorithmic}
\end{algorithm}